\documentclass[letterpaper, 10 pt, conference]{ieeeconf}  

\IEEEoverridecommandlockouts                              

\usepackage[margin=0.75in]{geometry}

\usepackage{framed}
\usepackage{caption}
\usepackage{graphicx}
\usepackage{float}
\usepackage{todonotes}
\usepackage{amsmath} 
\usepackage{amssymb}  
\usepackage{array}
\usepackage{hyperref}
\newcolumntype{C}[1]{>{\centering\let\newline\\\arraybackslash\hspace{0pt}}m{#1}}

\title{Balancing and Walking Using Full Dynamics LQR Control \\ With Contact Constraints}

 \author{Sean Mason$^{1}$, Nicholas Rotella$^1$, Stefan Schaal$^{1,2}$, and Ludovic Righetti$^{2}$
\thanks{This research was supported in part by National Science Foundation grants IIS-1205249, IIS-1017134, CNS-0960061, EECS-0926052, the
DARPA program on Autonomous Robotic Manipulation, the Office of Naval Research, the Okawa Foundation, the Max-Planck-Society and the European Research Council under the European Union’s Horizon 2020 research and innovation program (grant agreement No 637935). Any opinions, findings, and conclusions or recommendations expressed in this material are those of the author(s) and do not necessarily reflect the views of the funding organizations.}
\thanks{$^{1}$Computational Learning and Motor Control Lab, University of Southern California, Los Angeles, California.}
\thanks{$^{2}$Autonomous Motion Department, Max Planck Institute for Intelligent Systems, Tuebingen, Germany.}
}

\begin{document}

\maketitle
\thispagestyle{empty}
\pagestyle{empty}

\begin{abstract}
Torque control algorithms which consider robot dynamics and contact constraints are important for creating
dynamic behaviors for humanoids. As computational power increases, algorithms tend to also increase in complexity.
However, it is not clear how much complexity is really required to create controllers which exhibit good performance.
In this paper, we study the capabilities of a simple approach based on contact consistent LQR controllers designed around key poses
to control various tasks on a humanoid robot. We present extensive experimental results on a hydraulic, torque controlled humanoid performing balancing and stepping tasks. This feedback control approach captures the necessary synergies between the DoFs of the robot to guarantee good control performance.
We show that for the considered tasks, it is only necessary to re-linearize the dynamics of the robot at different contact configurations and that increasing the number of LQR controllers along desired trajectories does not improve performance. Our result suggest that very simple controllers can yield good performance competitive with current state of the art, but more complex, optimization-based whole-body controllers. A video of the experiments can be found at \url{https://youtu.be/5T08CNKV1hw}
\end{abstract}

\section{Introduction} 

Biped robots that are expected to locomote in human environments require whole-body controllers that can offer precise tracking and well-defined disturbance rejection behavior. Although walking is a complex task involving both hybrid dynamics and underactuation, the level of controller complexity required to execute such a task is unclear. In recent years, optimal control strategies have seen success both in simulation and on real systems for torque controlled humanoids. Previous work, \cite{koolenT,tedrakeR,fengS,bStephens2,aHerzog}, have utilized Quadratic Programs (QPs) to compute inverse dynamics control optimized over a variety of constraints (e.g. dynamic consistency, joint tracking, friction cones, etc.). Trajectories are often planned in operational space and then converted to joint torques using the QPs. The problem can further be organized into hierarchies to solve whole-body control problems according to a set priority in goals such that tasks of higher priority will always be achieved first \cite{kanounO}. Unfortunately, along with the growing flexibility of these methods comes added computational overhead, complexity in tuning, and a lack of theoretical disturbance rejection metrics (such as the gain and phase margin of classical controls).

In linear control theory, the infinite horizon linear quadratic regulator (LQR) controller is the optimal solution for tracking a steady state pose with quadratic cost on state error and control effort. LQR controllers have also been extensively used to locally stabilize non-linear systems. In our previous work \cite{mason2014}, we proposed a contact-consistent LQR control design for humanoid
robots. The advantage of the controller is that it explicitly takes into account the coupling between the different joints to create optimal feedback controllers for whole-body coordination. Additionally, this control policy is computationally lightweight and demonstrates robust push recovery behavior competitive with more sophisticated balance controllers which use QP solvers for inverse dynamics \cite{bStephens2,aHerzog}, rejecting impulses up to 11.7 Ns with peak forces of 650 N while in double support without stepping. Despite this good performance, previous work was limited to a single contact scenario using a single linearization of the dynamics.

More recently, \cite{posaM} proposed to use a similar approach to stabilize constrained systems in order to track dynamic behaviors. Feedback gains, although computed, are not directly executed. Instead, walking trajectories are stabilized by computing the cost-to-go of the time varying LQR problem. This cost is then used as the objective in an inverse dynamics controller, which is implemented using a QP solved at each control cycle.  This optimization-based approach allows for the use of additional constraints such as torque limits and contact friction cones but has a relatively high computational complexity compared to a simple LQR design. Further, only results in simulation were presented.

In this work, we extend our previous approach on contact-consistent LQR control and demonstrate that it can be used for more complex scenarios including switching contacts. In contrast to approaches which solve a QP at each control cycle, we explore the idea of using only a small number of LQR controllers computed from a set of predefined robot poses. Our hypothesis is that only a small number of LQR controllers corresponding to a contact-consistent linearization of the dynamics at key poses are necessary to stabilize complex tasks. We focus predominantly on real robot experiments: balancing under disturbances caused by upper body motion, single leg balancing, and switching between multiple linearizations and contact conditions for walking. Additionally, we experimentally study how the number of re-linearizations of the dynamics along a trajectory affects performance. As hypothesized, our experimental results suggest that very few LQR controllers around key poses (typically at different contact configuration) are sufficient to stabilize a wide variety of tasks, and increasing the number of controllers does not improve tracking. These results suggest that a simple set of LQR controllers with low computational complexity can be used to control a wide range of humanoid motions.

\section{Problem Formulation and Control Design}
This work extends the approach developed in \cite{mason2014}, where we derived a full state LQR using the full dynamics of the robot. The main difference with our previous work is the method used to enforce contact consistency. Previously we used a kinetic energy argument to project the dynamics, which was only valid for linearization at zero velocity states. The projection developed below is now valid around arbitrary states which was necessary for linearization along planned trajectories.

The equations of motion for a floating-base robot with $n$ total degrees of freedom (DoFs) including the floating base and $m$ contact constraints can be written as
\begin{equation}
\label{eq:dynamics}
 M(\theta) \ddot{\theta} + C(\theta,\dot{\theta})\dot{\theta}+ g(\theta) = S^T\tau +J_c^T(\theta) f_c,
\end{equation}

where the variables are defined as\vspace{-2mm}
\begin{table}[H]
\centering
\begin{tabular}{lll}
	$\theta \in \mathbb{R}^{n-6} \times SE(3)$ 							& : & generalized coordinates\\
	$M(\theta) \in \mathbb{R}^{n\times n}$ 			& : &inertia matrix\\
	$C(\theta,\dot{\theta}) \in \mathbb{R}^{n \times n }$ 		& : & centrifugal and Coriolis forces\\
	$g(\theta) \in \mathbb{R}^{n}$ 					& : & gravitational force\\
	$\tau \in \mathbb{R}^{n-6}$ 							& : & active joint torques\\
	$J_c \in \mathbb{R}^{m \times n}$ 					& : & contact Jacobian matrix\\
	$f_c \in \mathbb{R}^{m}$								& : & contact forces\\
	$S \in \mathbb{R}^{(n-6) \times n}$ 					& : & joint selection matrix
\end{tabular}
\vspace{-3mm}
\end{table}

Contact with the environment is represented through contact constraints on the endeffector as

\begin{equation}
\begin{bmatrix}
	J_c & \textbf{0}\\
	\dot{J_c} & J_c\\
\end{bmatrix}\begin{bmatrix}
	\dot{\theta}\\
	\ddot{\theta}
\end{bmatrix} = \begin{bmatrix}
	\textbf{0}\\
	\textbf{0}
\end{bmatrix}.
\label{eq:constraints}
\end{equation} 

Eq. \eqref{eq:constraints} enforces that an endeffector in contact must have zero velocity and zero acceleration. We numerically linearize the above dynamics to determine a linear time invariant (LTI) system of the form: 
\begin{equation}
\label{eq:SS}
\dot{x} = Ax +Bu.
\end{equation}
Section II of \cite{mason2014} contains a detailed discussion of the linearization process as well as how the dynamics are reformulated such that the contact forces are chosen to be consistent with the contact constraint. As mentioned in \cite{posaM}, the resulting linear system in uncontrollable. Because of this, and small errors due to numerical precision during the linearization process, the resulting matrix is commonly ill-condition. This make it difficult for off-the-shelf programs (e.g. minreal in Matlab) to numerically eliminate the uncontrollable states of the system, as required to solve the LQR problem. To resolve this we ensure that the constraints are embedded in the linearized dynamics by projecting the linearized system into the nullspace of the kinematic constraint of Eq. \eqref{eq:constraints} using the projection matrix $N$,
$$N = \text{null}\bigg(
 \begin{bmatrix}
	J_c & \textbf{0}\\
	\dot{J_c} & J_c\\
\end{bmatrix}\bigg).$$

Here, $N \in\mathbb{R}^{n \times (n-m) } $ is an orthonormal basis for the nullspace of the constraint equation which maps the linearized dynamics to the minimal system as follows. 

\begin{table}[H]
\centering
\begin{tabular}{lll}
$x_m$ &=& $N^Tx$ \\
$A_m$ &=& $N^TA N$\\
$B_m$ &=& $N^TB$\\
$R_m$ &=& $R$\\
$Q_m$ &=& $N^TQN$\\
\end{tabular}
\end{table}

This method of projecting the dynamics into the constrained subspace is more general than previously shown in \cite{mason2014} because it is mathematically valid for linearizing around non-static poses, whereas the previous approach was not. This process enables one to linearize around predefined poses or the current state of the robot at each point along a trajectory as for the time varying LQR problem. The optimal feedback gain matrix $K_m$ are computed by minimizing the following cost function in the reduced state space
\begin{equation}
\label{eq:cost}
J = \int_0^\infty (x_m^TQ_mx_m+u^TR_mu)dt
\end{equation}
We then map the gains from the minimal system, $K_m$, back to the full system as
$$ K = K_m N^T.$$
The resulting controller for the humanoid robot is thus
\begin{equation}
\tau = \tau_0 - Kx.
\end{equation}

Where $\tau_0$ is the vector of joint torques that compensate the dynamics of the robot around the linearized state \cite{mason2014}.
For example, for a linearization at zero velocity poses, it corresponds to a gravity compensation term.
\begin{figure*}[t]
\centering
\includegraphics[width= 6.95in]{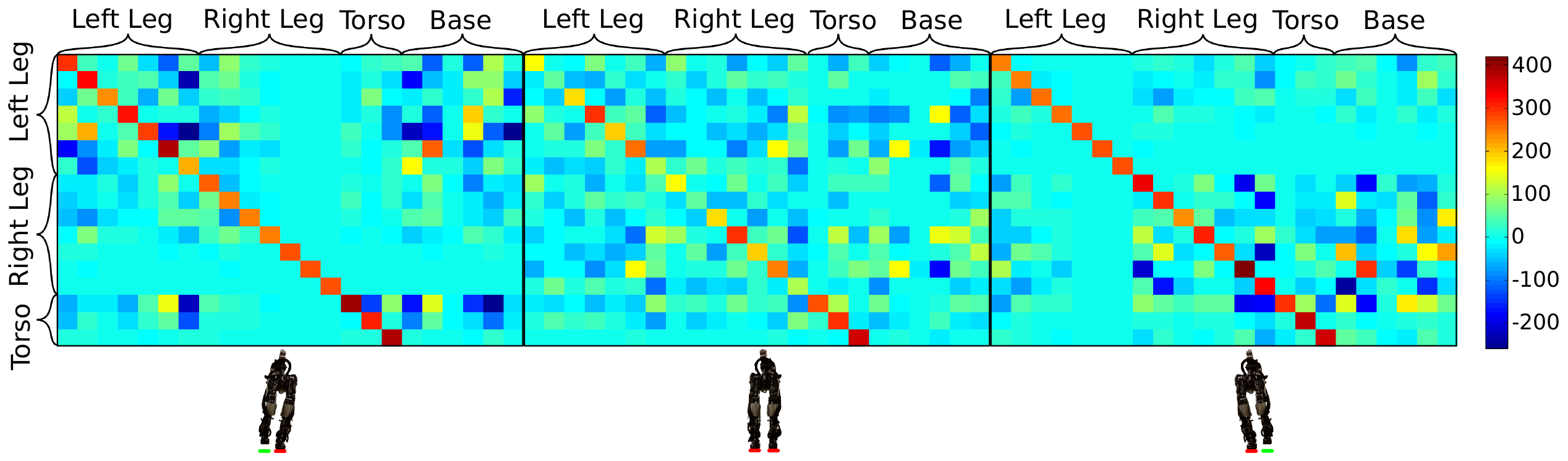}
\caption{Joint position portion of gain matrices produced from key poses and constraint conditions. The outer matrices are created for single leg balancing and the center matrix is double support balancing. Note that while in double support the LQR controller produces a highly coupled gain matrix (i.e. large off diagonal terms). In single support, the balancing strategy changes and the support leg is coupled among its own joints while mostly ignoring the tracking of the swing leg which becomes a diagonally dominated controller that mostly considers decoupled joint tracking.}
\label{fig:gain_matrix}
\end{figure*}

\section{Experiments}

In this section, we present experiments which evaluate the performance of the LQR control framework in a number of different scenarios. Section \ref{sec:disturbances} demonstrates the use of the controller to stabilize the robot during upper body disturbances.  Section \ref{sec:single_leg} presents results for robust disturbance rejection in single support.  Section \ref{sec:multiple_linearizations} investigates the number of LQR linearizations necessary to track a side-to-side motion.  Finally, sections \ref{sec:walking_simulation} and \ref{sec:walking_robot} present a framework and results for static walking control in simulation and on the real robot, respectively.

All robot experiments were conducted on the lower body of a hydraulic Sarcos humanoid using joint torque controllers as described in \cite{aHerzog} with slight modifications as discussed in Section \ref{sec:discussion}. At the joint position level, we combined moving average filtering onboard the motor controller cards (which run at 5Khz) with offboard second-order Butterworth filtering at 1Khz. The moving average filter used a window size of 16 measurements; the Butterworth filter had a cutoff frequency of 40 Hz.  This combination was determined empirically to remove noise without incurring a large delay, improving feedback control stability overall. All experiments used a foot size of 12.5 cm by 25 cm.  

\subsection{Linearizing Around Different Postures and Contacts} 
\begin{figure}[h]
\centering
\includegraphics[width = 8cm]{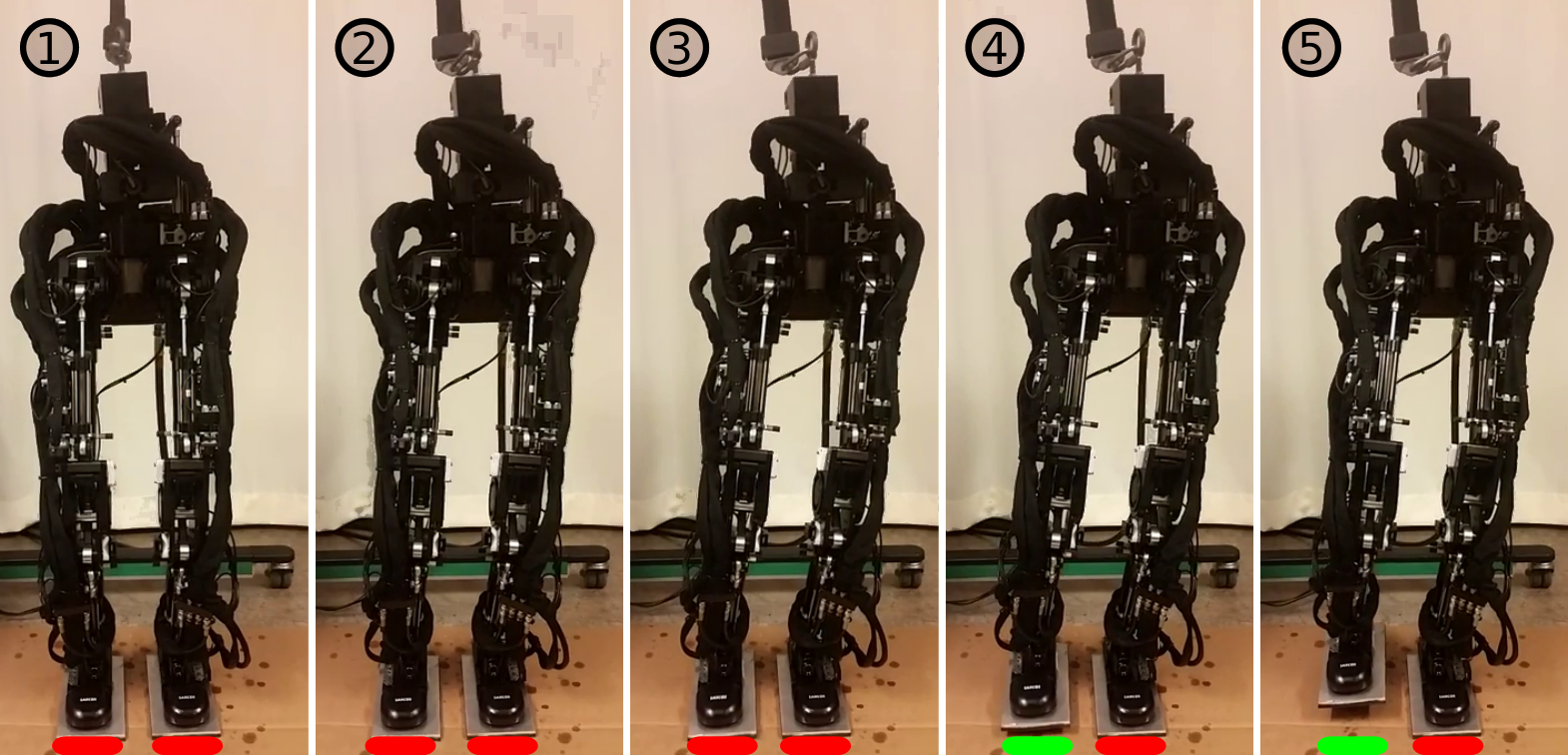}

\vspace{3mm}

\begin{tabular}{|C{1cm}|C{1cm}|C{1cm}|C{1cm}|C{1cm}|C{1cm}|}
\hline
	Pose \#& 1 & 2 & 3 & 4 & 5 \\ \hline
	Norm & 0 &  28.03 & 56.18 & 545.84 & 548.07\\ \hline
\end{tabular}

\caption{Postures that were chosen for linearization along with the Frobenius norm $\|K_n - K_1\|$ of the gain differences with respect to the first pose (base centered over feet).}
\label{fig:norm}
\end{figure}

By visualizing the resulting gain matrices from linearizing around different poses and contact conditions as shown in Fig. \ref{fig:gain_matrix}, we gain insight on the balancing strategy by inspecting the coupling between joint states and output torques. This type of analysis is not readily available for other controllers which output joint torques rather that a local feedback policy. High costs on the diagonals indicate decoupled joint tracking while off diagonal terms indicate coupling between different states (note that a perfectly diagonal matrix would correspond to a standard independent joint PD controller). Figure \ref{fig:norm} shows different poses and contact conditions used in this study and provides values of the Frobenius norm of the difference between gain matrices with respect to those for the centered posture. We can see that the gains change only slightly when shifting to the side compared to when the contact constraint changes, indicating that switching constraints has a much larger effect on the LQR solution. This insight is further explored experimentally in Section \ref{sec:multiple_linearizations}. In general, the diagonal terms produced by the LQR controller are lower than the independent joint PD controller typically used on the robot. The Frobenius norm of the LQR gain matrices used on the robot and independent joint PD matrix are approximately 650 and 1,500 respectively.

\subsection{Disturbances Caused By Upper Body Motion}\label{sec:disturbances}
\begin{figure}[h]
\centering
\includegraphics[width=2.95in]{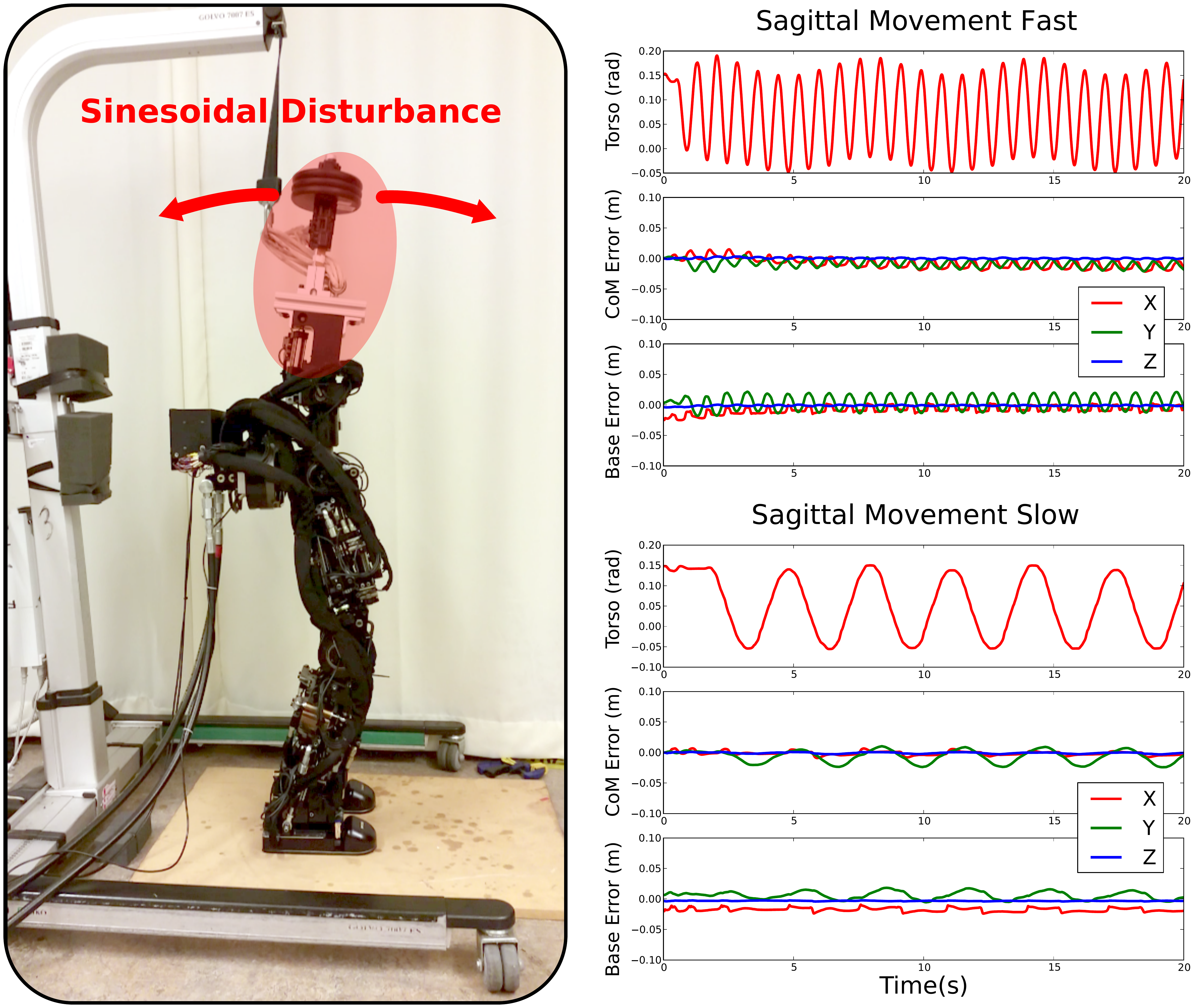}
\caption{The LQR controller was used to balance the lower body of the robot without knowing about the motion of the upper body. The graphs show the error in the base position for both a slow (0.2 Hz) and fast motion (0.8 Hz).}
\label{fig:balanceExperiment}
\end{figure}

In \cite{mason2014}, the LQR controller was evaluated through external pushes on the robot with impulses of up to 11.7 Ns along the sagittal plane. While disturbances are often external in nature, they can also originate from the motion of the robot itself. It is a common scenario for a humanoid to engage in tasks that decouple upper and lower body goals. The upper body may be moving around and interacting with the world while the lower body is purely focused on balancing. To simulate this, a mass of 10kg was added to the torso joint and moved through sinusoidal motions of different frequencies. The LQR controller was able to balance for motions of the upper body moving up to 0.8 Hz in the sagittal plane and 0.5 Hz in the frontal plane with an amplitude of 0.1 rad. Plots of the sagittal disturbance tests for slow and fast sinusoids are shown in Fig. \ref{fig:balanceExperiment}. 

\subsection{Single Leg Balancing}\label{sec:single_leg}
In the case of locomotion over very rough or complex terrain, stepping to maintain stability may not always be possible. An example of this is the problem of crossing stepping stones; there are a finite number of safe regions to step and the robot must be able to reject disturbances or correct for error while in single support so it can plan and execute safe stepping motions. In the following experiment, we test the ability of the LQR controller to balance in single support while being externally perturbed in different locations. Referring to the gain matrices in Fig. \ref{fig:gain_matrix}, one can see that the balancing strategy for single support is drastically different than that for double support. In the gain matrix, it is clear that the support leg has a large amount of coupling (shown by off diagonal terms) to help the robot balance, while the swing leg essentially becomes decoupled from the support leg (primarily diagonal terms) with some coupling relating the hip position to base error (i.e. the swing leg can be moved at the hip to help correct for errors measured in the base). This balancing strategy can be seen on the robot in \url{https://youtu.be/5T08CNKV1hw}. 

\begin{figure*}[t]
\centering
\includegraphics[width= 7in]{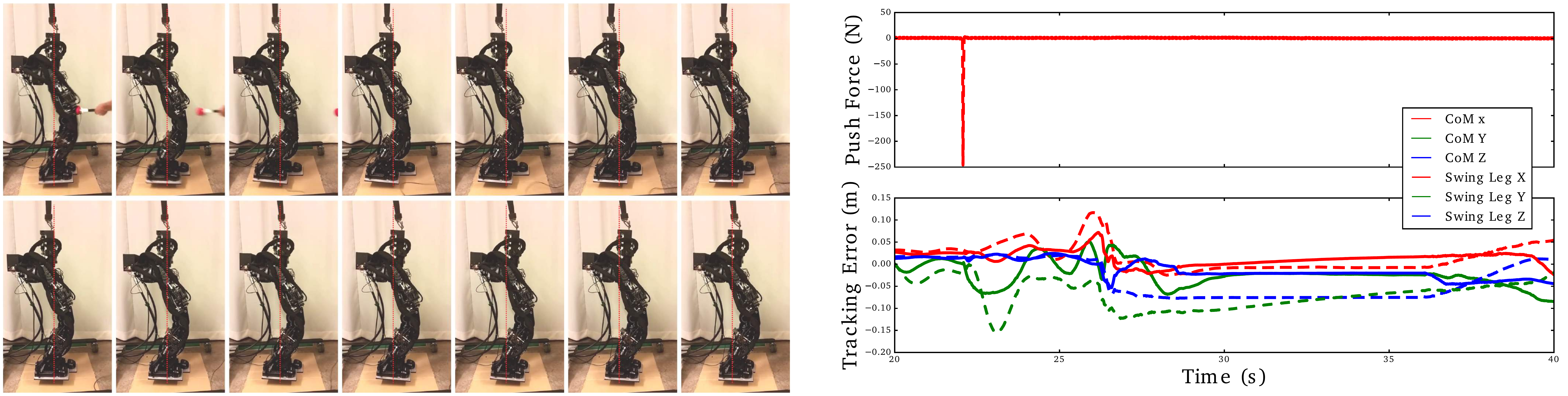}
\caption{Snapshots from single support balancing task and the tracking error of the CoM and swing leg after an external impulse. The impulse was 8.2 Ns with a peak force of 488 N. A similar test was conducted for a push located at the base of the robot, resulting in a impulse of 5.5 Ns and peak force of  246 N. This value is very close to the max impulse of 5.8 Ns measured when using a hierarchical QP controller on the same robot \cite{aHerzog}.}
\label{fig:ssp}
\end{figure*}

\subsection{Switching Between Multiple Linearizations}\label{sec:multiple_linearizations}

Intuitively, using more linearizations should better capture the dynamics of the robot.  However, the gain matrices produced by varying the pose changed very little compared to those produced by changing the contact constraints. To quantify the effect of using multiple linearizations, we tracked a side-to-side motion and transitioned between sets of gains, again using the control diagram shown in Fig. \ref{fig:block_diagram}. The root mean squared error (RMSE) for CoM tracking was calculated for two different motion speeds, as summarized in Table \ref{table:rmse}. 
\begin{table}[H]
\centering
\begin{tabular}{|c|c|c|c|c|c|}
\hline
	Poses & SSP (s) & DSP (s) & X (m)& Y (m)& Z (m)\\
\hline
	1 	&4&4& 0.00806 \,\,& 0.08239 \,\,& 0.30955 \,\,\\
\hline
	3 	&4&4& 0.00814 \,\,& 0.08268 \,\,& 0.30967 \,\,\\
\hline
	5 	&4&4& 0.00816 \,\,& 0.08262 \,\,& 0.30962 \,\,\\
\hline\hline
	1 	&0.5&2& 0.01287 \,\,& 0.02913 \,\,& 0.03900 \,\,\\
\hline
	3 	&0.5&2& 0.01304 \,\,& 0.05456 \,\,& 0.06337 \,\,\\
\hline
	5 	&0.5&2& 0.01299 \,\,& 0.028760 \,\,& 0.04142 \,\,\\
\hline
\end{tabular}
\caption{RMSE for tracking a side-to-side ZMP trajectory with no contact switching. The single support phase (SSP) time refers to the dwell of the ZMP located at a single foot, where the foot would be lifted if the robot were walking, and the double support phase (DSP) refers to the transition of the ZMP from one foot to the other. The number in the poses column indicates the number of linearized poses used for tracking.}
\label{table:rmse}
\end{table}
From the experiments, we saw no clear advantage to using a larger number of linearizations. We believe that this indicates that for the scenarios presented, that there are other bottlenecks on the real robot (such as sensor noise and torque tracking) that dominate the system's performance. We thus conclude that using multiple linearizations for a given contact state gives no advantage over using a single linearization on the real robot. 

\subsection{Walking: Multiple Linearizations and Contact Switching}\label{sec:walking}
\subsubsection{Simulation}\label{sec:walking_simulation}
While any walking planner could be used, we chose to generate walking trajectories offline. In doing so, we rely solely on the ability of the proposed LQR control architecture to robustly track the original planned trajectory. We used the preview control approach proposed by \cite{sKajita} based on a ZMP stability criterion. From the resulting CoM and predefined foot step trajectories, we generate desired joint space trajectories using inverse kinematics and feed-forward torques from gravity compensation. During double support, we exploit torque redundancy to optimize contact forces to shift the robot's weight between successive stance legs \cite{righetti2012}. We found that this weight distribution optimization is crucial in order to achieve the desired motion and ZMP. 

\begin{figure}[H]
\centering
\includegraphics[width=3 in]{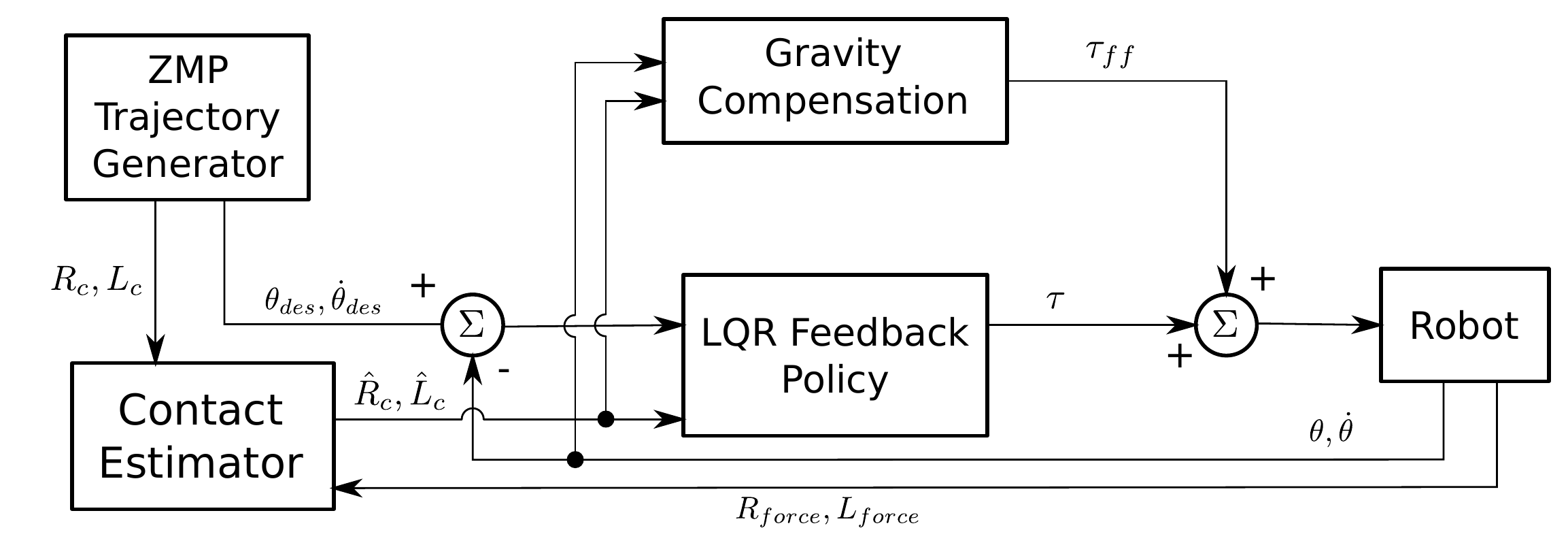}
\caption{Block diagram depicting the control architecture used in the walking experiments. $\theta$ is the joint and base state, $\tau_{ff}$ is the feed-forward torque, and $R_c$ , $L_c$ denote the time varying contact trajectories for the feet.}
\label{fig:block_diagram}
\end{figure}

From the preceding results, we know that the LQR formulation behaves well for tasks without contact switching. Switching contact conditions based solely on the planned trajectory often cause instability because the real contact with the environment never follows the exact timing of the planned trajectory. For the walking task, it thus proved essential to estimate the contact state using both the measured normal force and the planned trajectory. To ensure that contacts are created robustly during the swing leg touch down, trajectories were designed such that the foot contacted the ground with nonzero velocity. Furthermore, to deal with joint torque discontinuities at contact transitions, caused if one were to simply switch from one LQR controller to another, we use a heuristic approach to blend the gains produced by the LQR controllers. When the contact condition changed, we quickly interpolate to an independent joint PD controller from the previous LQR gain matrix and then interpolate to the new LQR gain matrix. The gain matrices were selected by considering the estimated contact condition of the endeffectors and the minimum norm of the current posture to the pre-selected linearized postures. The control scheme used for walking is shown Fig. \ref{fig:block_diagram}.

\begin{figure}[H]
 \centering
\includegraphics[width=3in]{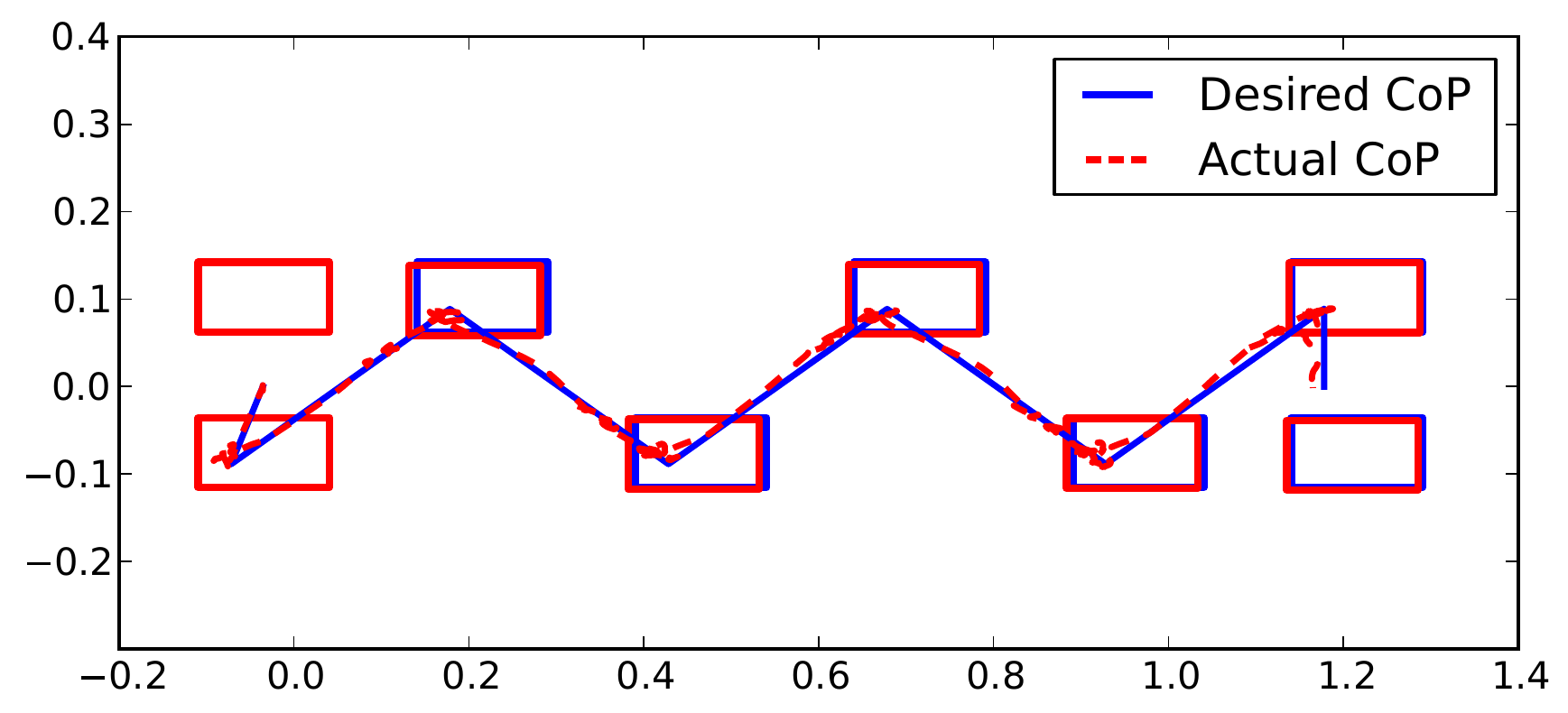}
\caption{CoP and footstep location tracking for walking using the LQR controller in simulation. }
\label{fig:walking_cop}
\end{figure}

In simulation, the robot could precisely track the prescribed ZMP trajectory and planned footsteps for a wide range of timings and step parameters using only three linearizations (one for double support and one for each single support pose) as shown in Fig. \ref{fig:walking_cop}. The close tracking of the open loop trajectory over a large number of steps demonstrates the effectiveness of using only a small number of linearizations.  
\begin{figure*}[t]
\centering
\includegraphics[width= 6.8in]{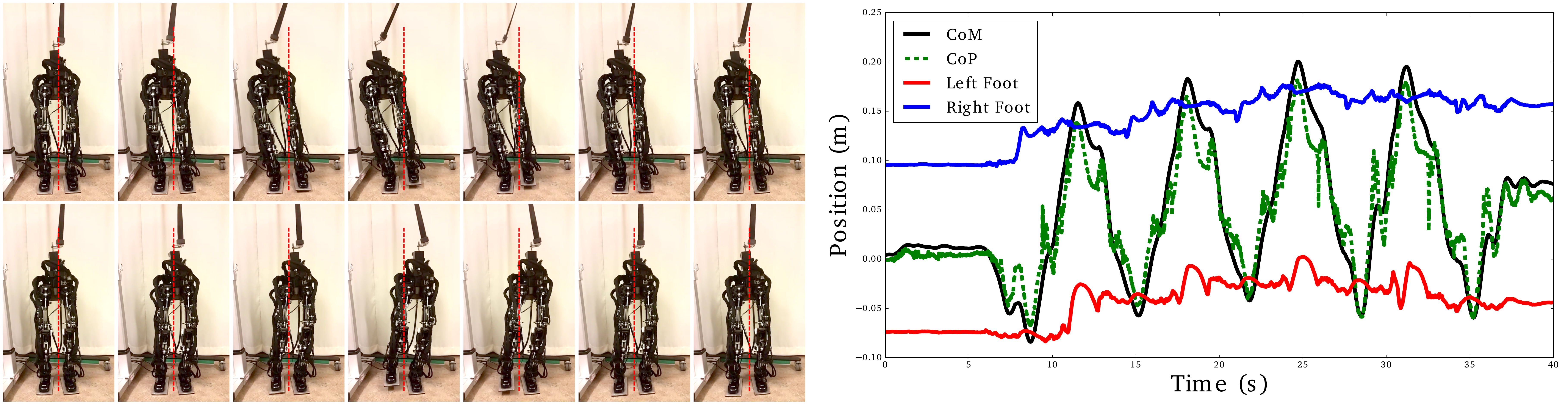}
\caption{Snapshots from the walking experiment on the real robot using multiple LQR linearizations and contact switching. On the right we show the CoP, CoM, and foot locations in the side-to-side direction throughout the experiment.}
\label{fig:walking}
\end{figure*}

\subsubsection{Real Robot}\label{sec:walking_robot}
Expanding on the simulation results, we tested the proposed control architecture on the real robot to walk-in-place. Fig. \ref{fig:walking} shows snapshots from the walking experiment as well as the CoP and CoM relative to the feet for a series of nine steps. The maximum walking speed we were able to achieve had a 1.5s SSP and a 1.8s DSP. As the robot walked at faster rates, the placement of of the foot steps became less precise and the performance degraded because of inconsistencies with the pre-planed trajectory. If the feet land in the wrong location, the desired pose would become kinematically infeasible and the LQR controller would be physically unable to drive the error to zero. We believe that replanning the trajectory online would alleviate this issue and allow for both faster and more stable walking.

\section{Discussion}\label{sec:discussion}
The experiments of the preceding section demonstrate both the strengths and shortcomings of the whole body LQR control approach for a number of different tasks. We showed in the experiments that a relatively small number of linearizations were necessary to control
stepping tasks and that increasing the number of linearization along the tracked trajectories did not necessarily improve performance
on the real robot. Our experiments suggest that linearizations are only necessary at different contact configurations
and that a complete gain schedule along planned trajectories, where gains change every control cycle, might not provide
additional robustness in real robot experiments. It is also important to notice that we could not achieve stepping
and balancing tasks by merely using an independent joint
PD controller and that taking into account joint coupling improves significantly performance.

Properly damping the system was one of the biggest limitations to obtain robust behaviors on the robot. LQR controllers
tend to create aggressive feedback on joint velocity which can create issues on the robot due to excessive noise in the
velocity signals. While this problem does not exist in simulations, it becomes quickly visible on the real robot.
While a combination of onboard and offboard filtering helped, we expect that using a model-based filter with additional sensors such as IMUs can allow to increase the control bandwidth \cite{rotella}. Using control design techniques that explicitly take into account measurement uncertainty might also help addressing this issue \cite{bPonton}, published in WAFR 2016.

Additionally, we found that a significant issue in achieving stable LQR control was the performance of the low level torque feedback control loop. In our previous work, we tuned aggressive low-level torque controllers using velocity compensation to eliminate natural actuator damping and ensure good force tracking. While this worked sufficiently in other inverse dynamics approaches \cite{aHerzog}, we found that a too aggressive velocity compensation gain in the low-level controller tended to destabilize the higher-level LQR controller. In our experiments, velocity compensation was removed and torque feedback gains were lowered. While torque tracking bandwidth was slightly reduced, this allowed to use higher feedback gains for the LQR controller and significantly improved performance. The trade-off between force tracking performance and higher level control performance is extensively discussed in Focchi et. al \cite{Focchi2016}.

Contact transitions were also difficult to control and accurate contact estimation was an important element of the control architecture. Indeed, gains change drastically when contact conditions changes and poor  estimation of this can lead to unstable behaviors (e.g. when switching to a double support control when the robot is still in single support). In addition, our heuristic used to smooth transitioning between gains at each contact sequence during the fast contact transition also significantly increased performance. While controlling contact transitions was not an issue in simulation, it remains an important issue for implementation on real robots. 

\section{Conclusion}

In this paper we formulated a time invariant LQR controller and show on a real robot that this computationally lightweight control policy can be used to combine a small number of linearizations to create complex motion. By using only a single linearization around a key pose projected into contact constraints, we were able to stabilize upper body motion for slow (0.2 Hz) and fast (0.8 Hz) motions and balance on one foot and while rejecting pushes of up to 8.2 Ns with peak forces of 433N without stepping. We then explored using multiple linearization to tracked side-to-side motions and found that on the real robot there was no measurable advantage for using a higher number of linearizations. Finally, using linearizations around each contact situation (double support and both single support poses) was sufficient enough to track a ZMP walking trajectory when coupled with a contact estimator that helped transition between contact switching. These results highlights both the ability to control complex motion with relatively simple control policies and also the need to evaluate modern algorithms on real hardware. On the robot, we observed that the aggressive stabilizing commands LQR produces could be problematic with high sensor noise and limited control bandwidth. In the future, we plan to incorporate joint state estimation \cite{rotella}, focus on robustifying contact transitions and test the control framework in combination with online model predictive control planners.

\bibliography{references}{}
\bibliographystyle{IEEEtran}
\end{document}